\documentclass[10pt,twocolumn,letterpaper]{article}

\usepackage{cvpr}
\usepackage{times}
\usepackage{epsfig}
\usepackage{graphicx}
\usepackage{amsmath}
\usepackage{amssymb}
\usepackage{caption}
\usepackage{subcaption}
\usepackage{wasysym}

\usepackage[pagebackref=true,breaklinks=true,letterpaper=true,colorlinks,bookmarks=false]{hyperref}

\usepackage{xcolor}

\graphicspath{{images/}}

\cvprfinalcopy 

\ifcvprfinal\fi
\begin{document}

\title{SPIN: A High Speed, High Resolution Vision Dataset for \\ Tracking and Action Recognition in Ping Pong }

\author{Steven Schwarcz\thanks{Authors contributed equally.}
\hspace{1mm}\thanks{Work done during internship at Robotics@Google.}\\
University of Maryland\\
College Park \\
{\tt\small schwarcz@umaics.umd.edu}
\and
Peng Xu\\
Robotics at Google\\
{\tt\small pengxu@google.com}
\and
David D'Ambrosio\\
Robotics at Google\\
{\tt\small ddambro@google.com}
\and
Juhana Kangaspunta\\
Robotics at Google\\
{\tt\small juhana@google.com}
\and
Anelia Angelova\\
Robotics at Google\\
{\tt\small anelia@google.com}
\and
Huong Phan\\
Robotics at Google\\
{\tt\small huongphan312@gmail.com}
\and
Navdeep Jaitly\textsuperscript{*}\\
Robotics at Google\\
{\tt\small ndjaitly@google.com}
}
\maketitle
\begin{abstract}
We introduce a new high resolution, high frame rate stereo video dataset, which we call SPIN, for tracking and action recognition in the game of ping pong. The corpus consists of ping pong play with three main annotation streams that can be used to learn tracking and action recognition models -- tracking of the ping pong ball and poses of humans in the videos and the spin of the ball being hit by humans. The training corpus consists of 53 hours of data with labels derived from previous models in a semi-supervised method. The testing corpus contains 1 hour of data with the same information, except that crowd compute was used to obtain human annotations of the ball position, from which ball spin has been derived. Along with the dataset we introduce several baseline models that were trained on this data. The models were specifically chosen to be able to perform inference at the same rate as the images are generated -- specifically 150 fps. We explore the advantages of multi-task training on this data, and also show interesting properties of ping pong ball trajectories that are derived from our observational data, rather than from prior physics models. To our knowledge this is the first large scale dataset of ping pong; we offer it to the community as a rich dataset that can be used for a large variety of machine learning and vision tasks such as tracking, pose estimation, semi-supervised and unsupervised learning and generative modeling.
\end{abstract}
\vspace{-8mm}
\section{Introduction}

\begin{figure}[t]
 \centering
 \begin{subfigure}[b]{0.23\textwidth}
     \centering
     \includegraphics[width=\textwidth]{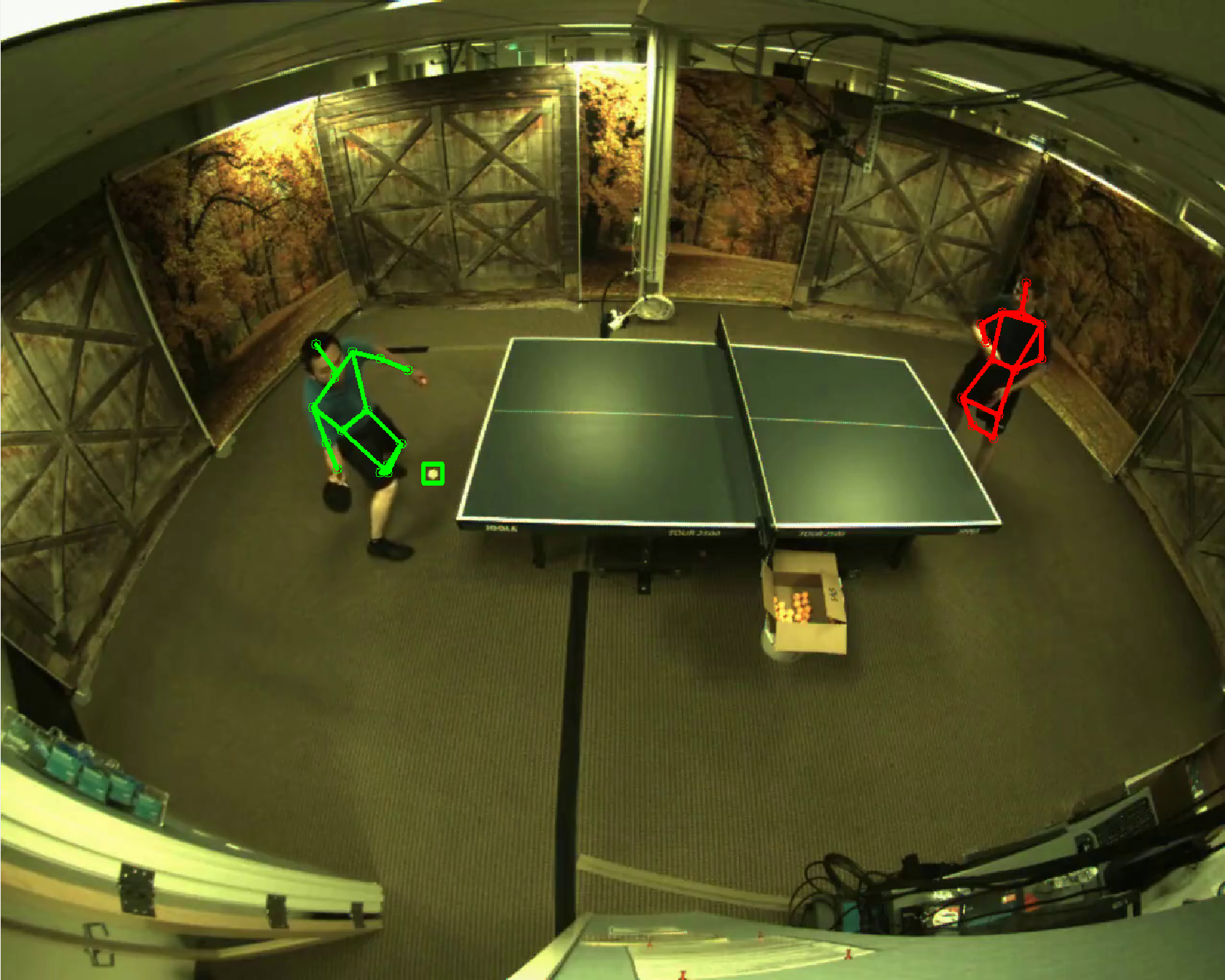}
 \end{subfigure}
 \begin{subfigure}[b]{0.23\textwidth}
     \centering
     \includegraphics[width=\textwidth]{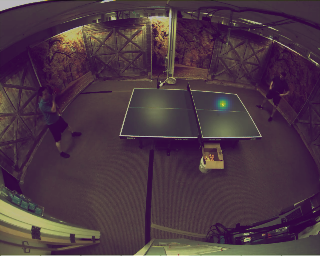}
 \end{subfigure}
\caption{We present a dataset for analyzing ball dynamics and human actions in the game of ping pong. Left: Ground truth annotations for a frame from the SPIN dataset. SPIN contains annotations for the ball location and human pose at every frame, as well as information about future frames, such as how much the next hit will put on the ball and where the ball will bounce after it is hit. Right: A heatmap for ball location, predicted by our method. }
\label{fig:intro}
\end{figure}

Sports datasets pose several interesting vision challenges ranging from object detection, tracking and action prediction and as such there has been significant interest in using sports datasets for training vision and other models. For example, basketball, ice-hockey and other sports have been actively explored in the past for building vision models \cite{ramanathan2016detecting, svw} and for generative models \cite{sun2019stochastic}. Most of these datasets use either low resolution images or slow frame rates for single cameras and focus on offline modeling. 

In this paper, we introduce a new high resolution, high frame rate stereo video dataset for tracking and action recognition in the game of ping pong, which we call SPIN. The high resolution, and high frame rate, stereo videos allows the possibility of inferring 3D trajectories of ping-pong balls and 2D human pose in the videos. This allows for more accurate predictive modeling, generative modeling, and even semi-supervised learning. Traditional vision models based on architectures such as ResNet are expected to work quite well on these datasets \cite{he2016deep}. 

However, when online inference of high frequency, high resolution images is required, traditional models simply cannot performed at the rate of image acquisition. In this paper, we only explore models that can perform inference online, at the speed of data generation.

The corpus we are releasing consists of ping pong play with three main annotation streams that can be used to learn tracking and action recognition models -- tracking of the ping pong ball, human poses and spin of the ball. The training corpus consists of 53 hours of data (about 7.5 million frames of active play) recorded from 2 cameras simultaneously with annotations derived from previously trained models. The testing corpus contains 1 hour of data with the same information, albeit annotated by humans. Note that spin information was not annotated directly, but rather computed from the ground truth ball positions.

Along with the SPIN dataset we introduce several baseline models that were trained on this data, including recurrent models (Conv-LSTMs, Conv-Gated-Sigmoid), and single frame non-recurrent models. The models were specifically chosen to be able to perform inference at the same rate as the images are generated, which is roughly 6.6 ms per stereo pair (150 fps) of RGB images of size 1024 x 1280 \footnote{We actually train most of our models using Bayer patterns directly to limit the time it takes to load images on to a GPU.}. We explore the advantages of multi-task training on this data, and also show interesting properties of ping pong ball trajectories that are derived from our observational data, rather than from prior physics models. For instance, we show that the dynamics of the ball's movement give rise to 3 distinct clusters, representing different levels of top spin applied to the ball.

In this paper we only focus on the above three tasks for visual prediction models. However, since the dataset we are releasing is a high speed, high resolution stereo dataset, much richer information can be derived from the same. For example, the dataset can be used for 3-D human pose tracking, and for richer action prediction tasks such as forehand shots, back-hand shots etc derived from the pose information. A multi-task dataset such as this could benefit from powerful vision models, such as those using attention, and the dataset can also be used for semi-supervised and unsupervised learning. We offer the dataset to the vision research community with the hope of spurring on interesting explorations in these and other related tasks.

\section{Related Works}
In this section we discuss the relationship between the models we used for baselines on our tasks and other related models. The main tasks we looked at were detection / tracking (of the ping pong ball), pose prediction (of human pose) and action recognition (spin prediction).

\subsection{Action Recognition}
One set of problems related to the task at hand is that of video action recognition. There are a large variety of datasets for this task approaching action recognition from different angles. This ranges from small, trimmed video datasets like UCF-101 \cite{Soomro2012UCF101AD} or JHMDB \cite{Jhuang2013}, to much larger datasets like Kinetics \cite{2017QuoVA} or Moments \cite{Monfort2018MomentsIT}. Other action recognition datasets, like Charades \cite{Sigurdsson2016HollywoodIH} or ActivityNet \cite{caba2015activitynet}, take longer form videos and make many temporal predictions in an untrimmed setting. Still other datasets, such as AVA \cite{Gu2017AVAAV} or ActEV \cite{Gleason2018APS}, take this even farther and require actions to be localized both spatially and temporally. When viewed as an action recognition dataset, SPIN is distinct from these, in that the focus is on predicting the future outcome of a player's current actions, as opposed to the action itself. Though we perform our experiments with SPIN in a trimmed context, the data we release is open to be explored in other contexts as well.

A variety of methods have been designed for these datasets \cite{Feichtenhofer2018SlowFastNF,tran2018}. One such method, i3D, fuses two streams of 3D convolutions; one taking in raw RGB frames and the other taking computed optical flow \cite{2017QuoVA}. A variety of methods extend from this architecture, for example by attempting to make optical flow computation more efficient \cite{Sun2017OpticalFG,Piergiovanni2018RepresentationFF,Diba2019DynamoNetDA}, or augmenting the architecture with attention based features \cite{Wang2017NonlocalNN}.

The sequential nature of videos creates an opportunity to make use of many techniques originally designed for language modelling. In this vein, many action recognition models make use of self-attention and similar techniques to achieve better performance \cite{a2nets}. This includes techniques such as second-order pooling \cite{Girdhar2017AttentionalPF,Long2017AttentionCP}, or generalizations of the transformer attention architecture \cite{Vaswani2017AttentionIA,Girdhar2018VideoAT} to visual data. Still other techniques make use of recurrent models \cite{Sharma2015ActionRU} or other gated structures \cite{Miech2017LearnablePW,Xie2017RethinkingSF}.

\subsection{Tracking}
Visual object tracking models have received signification attention in the community because of the crucial importance of tracking to downstream tasks such as scene understanding, robotics, etc. Tracking problems can be defined along a variety of directions\cite{fiaz2018tracking,luo2014multiple}. For example, tracking by detection relies on detecting a fixed subset of objects and then tracking them through videos sequences \cite{luo2014multiple}. On the other hand detection free trackers can deal with arbitrary objects that are highlighted at the start of the videos as being objects of interest \cite{held2016learning,feichtenhofer2017detect}. Evaluation for tracking is a further complicated issue, and can be formulated in various ways, such as whether new objects can appear or disappear in the videos and how resets are measured in assessing the performance of the trackers. In our dataset, the subjects used multiple different ping pong balls (sometimes using two at the same time), and hence it is appropriate to think of this as a multi-instance video detection problem, with appearance of new instances and disappearance of old ones. To create the dataset we performed tracking by detection by building a model for ping pong ball detection in videos and using dynamics models to produce smoother trajectories. However, the neural models we developed are technically video detection models and do not perform additional explicit tracking across multiple frames, such as is done by approaches such as \cite{feichtenhofer2017detect}. Nevertheless, these detection models can be incorporated into multi-object trackers that follow a more traditional tracking approach of associating detections that are close in a spatio-temporal sense and paired with approaches that allow for the appearance and disappearance of new tracks, such as that done by \cite{xiang2015learning} using reinforcement learning techniques.

\subsection{Dynamics of Ping Pong Balls}
Dynamics models that describe the motion of ping pong balls through the air, or across table and paddle bounces, are surprisingly intricate. The modelling of the aerodynamic forces acting on a flying spherical object in a game setting (\eg golf, ping pong, baseball, etc) has been a classical topic among the physics community \cite{briggs1959effect, mehta1985aerodynamics}. Yet the effects of aerodynamic drag and the Magnus effect, which causes ping pong balls to curve, are still being actively researched  \cite{robinson2013motion}.
The bounces of a ping pong ball are trivial to model when the relative motion between the ball and the surface is simplistic 
\cite{nagurka2003aerodynamic}, but quickly become complicated when the angle of contact and the ball's spin become variables \cite{widenhorn2016physics}.

One caveat is that there is no clear consensus on these modelling efforts \cite{chen2010dynamic}.
Another common theme of these works is a lack of data from real game conditions: experiments are usually performed in controlled setups such as wind tunnels where variables can be isolated, which does not cover all of the variations in ball velocity and spin that would exist in a real game. 

\section{Tasks}

\subsection{Ball Tracking}

Our new dataset affords the ability to explore many new tasks. The most fundamental of these tasks is ball tracking. For this task, we want to be able to identify the 2D image location of the ping pong ball as it moves through the air. Although we will consider this as a standalone task, it is so fundamental that we will also be learning it as a supplementary task in all of our other tasks.

\subsection{Spin Prediction}

When a highly skilled ping pong player hits a ball, they can often employ a technique to give the ball a topspin while it flies through the air. We empirically found that this gave rise to three different types of hit: ``no spin", ``light topspin", and ``heavy topspin," each with significant differences in aerial dynamics. In theory, there could exist another cluster for ``backspin" that causes the ball to fall slower, but in practice our dataset of competitive play did not include enough examples to actually parse out this cluster. Although we could train a model to learn which type of dynamics a ball has while it is flying in the air, this task would not be very challenging and follow naturally from a model that could effectively track a ball as it moved. 

Thus, rather than determine the current cluster of the ball, we set the task of trying to predict the future spin type cluster of the ball - the value it will have only once the player has actually hit it. This task relies on more than simple knowledge of the ball's location; a successful model must look at and understand the movements of the people who are playing the game.

\subsection{Pose Detection}

In addition to tracking ball location, we also explore the task of tracking human pose in 2D. This task is introduced primarily as a means for improving the results of spin classification; because spin prediction is a human-centric task (we are trying to determine what action the human is most likely to do), it is logical that providing additional signal about the humans movements and location could improve performance.
\section{Architectures}
\subsection{Tracking} \label{sec:gated}
To perform these tasks, we propose a fully-convolutional recurrent architecture. Our recurrent architecture is inspired by standard RNN variants suchs as Gated Recurrent Units (GRUs) \cite{Cho2014LearningPR} and LSTMs \cite{lstm,shi2015convlstm}, however we have simplified it to use a single stream of hidden states that are updated at each step. Given a previous hidden state $h_{t-1}$ and an input $x$, we perform our recurrent step as follows:

\begin{align}
z_t & = \sigma(W_z * h_{t - 1}) \\
c_t & = \text{tanh}(W_c * h_{t - 1}) \\
h_t & = z_t \astrosun c_t
\end{align}

\begin{figure}
     \centering
     \begin{subfigure}[b]{0.4\textwidth}
         \centering
         \includegraphics[width=\textwidth]{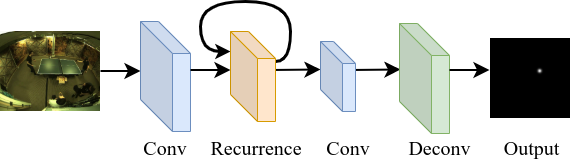}
         \caption{Tracking architecture}
         \label{fig:ball_arch}
     \end{subfigure}
     \begin{subfigure}[b]{0.4\textwidth}
         \centering
         \includegraphics[width=\textwidth]{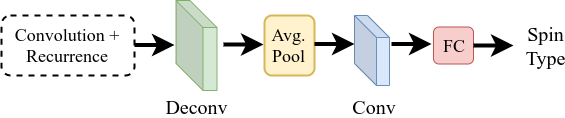}
         \caption{Spin classification architecture}
         \label{fig:spin_arch}
     \end{subfigure}
        \caption{A high-level view of the architectures we use. (a) The architecture we use for tracking uses convolution blocks to reduce resolution, a recurrent layer, and deconvolution layers to generate heatmaps for tracking. (b) When learning spin type, we branch off after the deconvolution before the last layer, sequentially performing average pooling, convolution, and flattening. We finish with a fully connected layer to predict the final spin type.}
\end{figure}

where $h_t$ is both the output and the new hidden state, $*$ represents application of a convolutional filter, $\sigma$ is a sigmoid function, and $\astrosun$ represents element-wise multiplication. This simplified architecture is similar to the one used by Oord \etal in \cite{Oord2016WaveNetAG}, making use of the gating mechanism of LSTMs or GRUs, while being simpler to implement and easy to train. 

Additionally, each step operation is performed using convolution, instead of fully connected components, which allows spatial information to propagate through the recurrent portions of the network.

To perform ball tracking, we feed in each $1280 \times 1080$ frame at full resolution, quickly reducing dimensions using a series of convolutions. We perform recurrence on these reduced features, increase resolution using deconvolution layers, and ultimately generate a  single channel $160 \times 128$ heatmap over the image by performing softmax over all spatial locations. The resolution of this heatmap, is less than that of the original image, so in order to maximize pixel-level accuracy we also learn a patch-based detection model, which extracts the features in a window around the maximal detection and feeds them into a smaller network that uses deconvolutions to upsample the features into a heatmap that more precisely locates the ball within the patch. Figure \ref{fig:ball_arch} describes our main architecture for generating the heatmaps, and the supplemental material contains specifics for both the ball detection architecture and the patch model mentioned above.. We train each heatmap using standard cross-entropy loss.

\subsection{Spin Type}

To learn which type of spin we would like the ball to have, we create a separate branch of the network to produce classification results. As shown in Figure \ref{fig:spin_arch}, we split off a new branch of our neural network after the recurrent layer. In order to keep the number of parameters tractable, the resolution of the feature maps was reduced with an average pooling layer. We then pass the reduced features through convolutional layers, flatten them, and pass them to two dense layers to create predictions that we train with cross-entropy loss.

\subsection{Pose}

We perform pose prediction similarly to ball tracking, producing 30 heatmaps (15 joints for 2 people) at each frame in the same manner as the heatmaps for the ball location. We learn these prediction targets as heatmaps in the final layer, learned with the same cross entropy loss used for ball tracking.



\section{Data}

\begin{figure}[t]
\begin{center}
\includegraphics[width=0.9\linewidth]{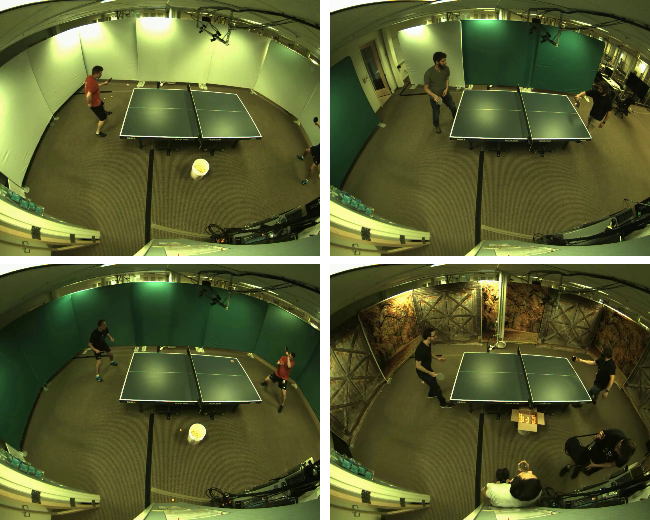}
\end{center}
\caption{Sample frames from the SPIN dataset.}
\label{fig:data}
\end{figure}

For all experiments performed here, we use a right-handed 3D coordinate system where $x$ and $y$ are parallel to the ground plane, and $z$ is vertical. The players themselves stand along the $y$ axis, with one player on the positive side and the other one on the negative side.

All of the visual data present in the SPIN dataset is recorded in high speed at 150 frames per second and an image resolution of $1024 \times 1280$. In order to create variety within the dataset, the walls behind the players are varied throughout the dataset, as can be seen in \ref{fig:data}. In particular, the eval dataset does not use any of the backgrounds present in the training set. 

\subsection{2D Tracking}
The original tracking data for the SPIN dataset is derived from a sequence of semi-supervised labelling of objects in the datasets. At the first stage, a simple color detector was used, along with an OpenCV \cite{opencv_library} mixture of gaussians background detection algorithm to generate a candidate of moving ping pong balls in each frame of the stereo image pairs. Only stereo-pair candidates with low reprojection errors were kept for each frame. The remaining candidates were then extended from frame to frame using Kalman Filters both in 3D coordinates of the stereo pairs and 2D coordinates of the images of the stereo pair. Trajectories found in this way were used to generate training data for the next detection model, which was then used to generate the next round of 3D tracking results, and 2D tracking targets. Each subsequent round of semi-supervised data generation qualitatively improved the 2D detector over video images. A single detector was trained for all cameras. The detectors were trained with strong data augmentation with hue, saturation noise, and image flips (with the corresponding flipping of target locations). 

\subsection{Computing Trajectories} \label{sec:compute_traj}

Once we have initial frame-by-frame ball detections from at least 2 cameras, we use stereo matching between the cameras to generate possible 3D coordinates for the ball at each frame. To create more consistent trajectories from these detections, we first attempt to discover two different types of  inflection points: "bounce inflections" where the ball bounces off the table and "return inflections" where the ball is hit by a player. 

To find bounce inflections, we sweep a sliding window over the detections, taking 6 at a time and fitting each set to a polynomial that is first order in $x$-$y$ plane (parallel to the ground) and second order in the vertical $z$ plane. From this polynomial we derive the velocity of the ball at the end of the window. We then compute the approximate velocity of the ball at the next 6 points, and record all of the locations where the signs of the vertical velocities differ. Finally, we perform non-maxima suppression to filter out any bounces that do not occur on the $x$-$y$ bounds of the table, and mark all remaining inflections as bounce inflections. 

When computing bounce inflections, we also keep track of the y velocity.  Thus, though we compute return inflections using the same velocity-checking algorithm we used for bounce inflections (albeit along the $y$ axis instead of the $z$), we enforce that they can only occur between 2 bounce inflections that have opposite $y$ velocities. This helps drastically reduce false positives. This is also causes hits that go out of bounds not to be detected, which for our purposes we did not consider problematic, as we wanted to focus our analysis on successful hits. If this is undesirable for another application, the procedure can easily be modified to make an exception for balls that go out of bounds by simply treating the end of the trajectory as a possible inflection point.

Finally, given a set of inflections, we smooth our final estimates using a form of statistical bootstrapping wherein we pass over the area between inflections with a sliding window of 20 points. In each window, we take 25 samples of 6 points, fit a polynomial to them, and average out the results to achieve our final estimate.

These trajectories provide a rich source of data. The models we discuss focus on predicting what type of hit the player will perform, but the trajectories are in 3D coordinates they could also be used to learn other interesting tasks. For instance, one could use our computed trajectory information to predict where on the table a ball will land after it is hit.

\subsection{Measuring Spin}
\begin{figure}
     \centering
     \begin{subfigure}[b]{0.235\textwidth}
         \centering
         \includegraphics[width=\textwidth]{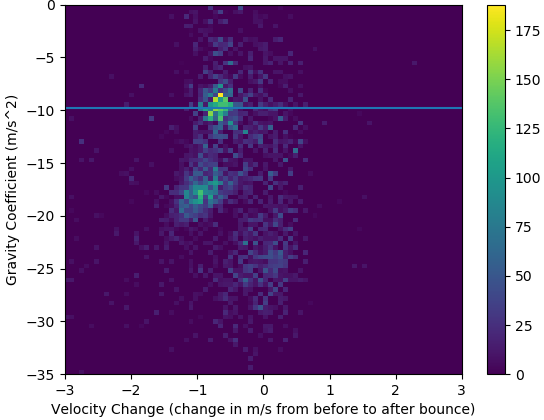}
         \caption{All players}
     \end{subfigure}
     \begin{subfigure}[b]{0.235\textwidth}
         \centering
         \includegraphics[width=\textwidth]{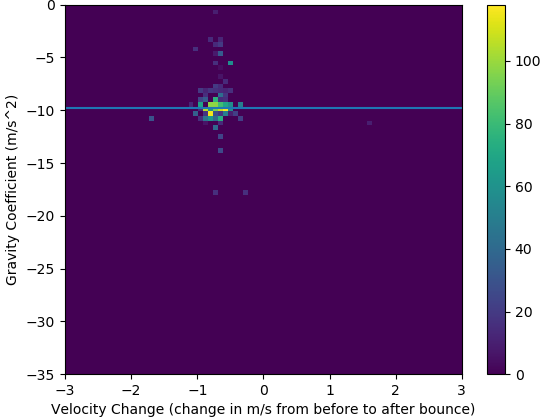}
         \caption{Amateurs}
     \end{subfigure}
     \begin{subfigure}[b]{0.235\textwidth}
         \centering
         \includegraphics[width=\textwidth]{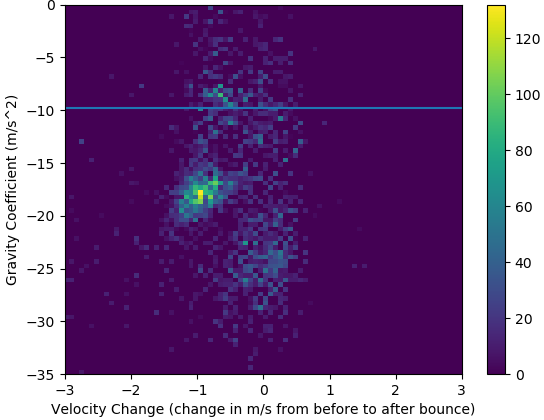}
         \caption{Professionals}
     \end{subfigure}
        \caption{A plot of downward acceleration against change in velocity for balls being hit during a game of ping pong. Note the three distinct clusters, only one of which exists when non-professionals play the game.}
        \label{fig:spin}
\end{figure}

One aspect of player actions that we wish to capture is the amount of spin that is placed on the ball when the player hit it. In order to do this, we show that it is possible to use information gleaned from ball trajectories to cluster different types of hits, as seen in Figure \ref{fig:spin}.

Given a ball trajectory immediately after a player hits the ball, we compute the change in x-y-velocity immediately before and after the ball bounces on the table, taken by measuring the movement 10 frames before and 10 frames after impact. Additionally, we compute the downward acceleration of the ball as it moves through the air by fitting a second-degree polynomial to the z coordinate of the ball through time. In other words, we having approximated the $z$ movement with a polynomial $ax^2 + bx + c$, Figure \ref{fig:spin} plots $2a$ on the $z$ axis. To allow the trajectory to better approximate a parabola, we remove the first and last 5 frames of the trajectory.

After plotting these results, we observe three clusters, as in Figure \ref{fig:spin}. When we divide the hits by “professional” and “non-professional” players, we notice that balls hit by non-professionals are extremely consistent - they fall at a steady $9.8 \frac{m}{s^2}$ and their hits have a very consistent change in velocity. When professionals play, we see two new clusters emerge. Through observation, we determined that these clusters correspond to topspin placed on the ball. The extra spin causes the ball to accelerate when it hits the table, causing a sharper change in velocity when hitting the table. Furthermore, in light of the Magnus effect, balls with topspin actually accelerate towards the table faster. 

From observation and discussions with the players, we determined that the two topspin clusters correspond to two distinct scenarios - the bottom cluster occurs when a player offensively hits the ball hard, and the middle cluster occurs when a player defends against an opponent's offensive hit. In the case of the middle cluster, the ball is still returned with a certain degree of topspin, but there is considerably less and thus the downward acceleration and coefficient of restitution are both reduced.

\begin{table}[]
\begin{tabular}{l | llll}
            & \begin{tabular}[c]{@{}l@{}}\textbf{Top}\\ \textbf{Cluster}\end{tabular} & \begin{tabular}[c]{@{}l@{}}\textbf{Middle}\\ \textbf{Cluster}\end{tabular} & \begin{tabular}[c]{@{}l@{}}\textbf{Bottom}\\ \textbf{Cluster}\end{tabular} & \begin{tabular}[c]{@{}l@{}}\textbf{No}\\ \textbf{Cluster}\end{tabular} \\ \hline
Train & 10173                                                 & 13849                                                    & 7317                                                     & 8456      \\
Eval  & 413 & 832 & 228 & 412
\end{tabular}
\caption{The number of hits in each cluster within the ``SPIN-OR" data split described in Section \ref{sec:spin_data}.}
\label{tbl:spin_counts}
\end{table}

Using these plots, we manually select the approximate centroids of these clusters located at $[-0.64, -9.5], [-0.9, -17.5]$ and  $[0.016, -24]$. When we process data for training, we assign each data point into a cluster based on which centroid it is closest to, and use this to assign a "spin type" to the hit. Table \ref{tbl:spin_counts} shows the number of hits assigned to each of the three clusters.

\subsection{Tracking Data} \label{sec:track_data}

Once we've computed and smoothed the trajectories as described in Section \ref{sec:compute_traj}, we create a training dataset out of the final trajectories. To do this, we use a stride of 15 and sample sets of 30 consecutive frames from the trajectory, meaning each frame of the trajectory is sampled twice. The ball detection is projected back into the image from its 3D coordinates, to be used as ground truth during training. Using this method, we generate 251,922 individual training examples over 212 videos (approximately 53 hours of game play) and 14,210 test examples over 13 videos. See Figure \ref{fig:data} for sample data. We refer to this dataset as the ``SPIN-All" dataset split, in contrast to the ``SPIN-OR" dataset split described below in Section \ref{sec:spin_data}.

\subsubsection{Human Annotation}

While our automatically generated ground truth is reasonably high quality, it is far from perfect. In order to properly measure the quality of our trained models, we also generate a set of human-annotated samples of ball trajectories, which we use for evaluation. For human annotation we chose to use a cloud-based crowd computation service that makes use of a pool of human workers. 

For annotation, we selected 11 three minute time segments of play that were collected with different backgrounds from training and with as much player diversity as possible. Both left and right camera images from the same time stamps were sent for annotation in order to allow association between cameras and stereo depth inference later on. The three minute clips of both cameras were further split into questions containing 200 frames each from a single camera. The 200-frames annotation was the unit of work for the human annotators. In order to associate tracks between questions, each question had a 10 frame overlap both with its temporal predecessor and successor to make it easier to associate tracks between consecutive questions.

The task for the workers was to annotate each track of a ball that was in play. A ball enters play when it is tossed up for a serve and exits play when it is no longer playable by a human player, e.g. on the floor or on the table. Each in-play ball received its own track. The tracks were comprised of sequential image-frame (x, y) center positions of the ball in each frame it was observed. Additionally, each center position is optionally annotated with a "Bounce" or "Hit" tag, indicating whether the ball bounced on the table or was hit by a player.

Before human annotation, the questions were pre-annotated by a ball tracking model. These annotations were used to minimize the time the humans would have to spend annotating each question.

\subsection{Spin Type Data} \label{sec:spin_data}

To predict spin type, we make some changes to how we prepare the data, creating an alternate dataset split we refer to as ``SPIN-OR" (for ``Only Return"). First, we limit our examples to those in which the ball is moving from left to right at the beginning of the example. Since our focus is primarily on the human who is preparing to hit the ball, only considering examples where the ball moves in a single direction allows the network to focus on a single individual, instead of needing to switch between individuals based on which direction the ball is moving. In order to avoid limiting the data unnecessarily, when the ball is moving right to left we flip the frame horizontally and reorder the human joints as necessary to make sure the example remains valid. Although one of our recorded players is already ambidextrous, using this flipped data has the added advantage of making the system robust to left-handed and right-handed play.

Next, we modify which frames we use. Instead of using all frames of play, we limit our examples to 25 frames surrounding the moment the ball is hit - 20 before the player hits the ball and 5 afterwards. We do this because we found empirically that using too many frames before the hit did not provide any additional useful signal - when the ball is on the opposite side of the table, it's too hard for the system to predict what the player will do. With this modified data generation, we create a dataset of 39,795 training samples and 1885 eval samples over the same 212 training and 13 eval videos used in Section \ref{sec:track_data}.

\subsection{2D Human Pose Data}

We also supplement all of our data in both splits with annotations for human pose. To generate these annotations, we pass our data through the pose detector described in \cite{Papandreou2018PersonLabPP}. This produces for each individual 15 detections per frame corresponding to 15 different human joint locations. In most cases, there are only two detections, corresponding to the two players in the game, however on rare occasions a bystander enters the frame and is detected. In these cases, we limit our annotations to only the two highest confidence detections. We further differentiate between the left and right player using the locations of their skeleton's detected bounding boxes, and add the final joint annotations to every frame in the data.
\section{Experiments}

\subsection{Implementation}\label{sec:impl}

For all experiments in this section we train on the short video snippets described in Section \ref{sec:track_data}, with 25 continuous frames per sample using a batch size of 8.  We train for over 100K steps starting with a learning rate of $0.002$ for the first thousand steps, then reducing to $0.00025$ for the rest of training. We also change the optimizer over the course of training, starting with stochastic gradient descent for the first $5000$ steps and switching to Adam \cite{Kingma2014AdamAM} afterwards for the remainder of training.

Although our datasets were built with 30 continuous frames per sample, we only use the first 25 for all experiments in order to reduce memory requirements. For the ``SPIN-OR" data split - used in the spin classification experiments - this means that our experiments are conducted with samples that contain 20 frames leading up to the hit, and 5 frames after the hit occurs.

Additionally, when training for spin classification, we only use videos of balls going left to right. Instead of throwing out samples with balls going the other way, we flip the image horizontally and permute the joints of labeled humans so that the human pose orientations remain consistent (\eg left ankle becomes right ankle). When learning on the ``SPIN-ALL" dataset split, we randomly flip images to provide additional signal. Finally, some experiments below are performed on a split we label ``OR + ALL", wherein we combine the ``SPIN-OR" and ``SPIN-ALL" dataset splits by sampling in equal weights from each. 

\subsection{Ball Detection}

\begin{table}[]
\centering
\begin{tabular}{l|c|c|c}
\textbf{Architecture}      & \textbf{AUC @ 2} & \textbf{AUC @ 5} \\ \hline
Gating (Proposed)      &   81.5   & 96.7              \\ 
LSTM                       &    80.6  & 93.0                    \\ 
Single Frame               &        88.5       & 96.0           \\ 
\end{tabular}        
\caption{Ball Tracking accuracy using various architectures showing, respectively, AUC at distance 2 and distance 5.}
\label{tbl:architectures}
\end{table}


\begin{table*}[]
\centering
\begin{tabular}{l| c c c c c c}
\textbf{Experiment} & \textbf{Spin Acc.} & \textbf{\begin{tabular}[c]{@{}l@{}}Pose \\ AUC@16\end{tabular}} & \textbf{\begin{tabular}[c]{@{}l@{}}Pose \\ AUC@40\end{tabular}} & \textbf{\begin{tabular}[c]{@{}l@{}}Ball \\ AUC@2\end{tabular}} & \textbf{\begin{tabular}[c]{@{}l@{}}Ball \\ AUC@5\end{tabular}} & \textbf{Dataset} \\ \hline
Spin                &        61.6           & -                                                                   & -                                                                  & -                                                                  & -                                                                 & OR               \\
Spin + Pose         &           68.2        &                                          77.2                          &                     91.5                                             & -                                                                  & -                                                                 & OR               \\
Spin + Ball         &         59.4         & -                                                                   & -                                                                  &             81.4                                                       &               94.6                                                    & OR + ALL         \\
Spin + Pose + Ball  &        72.8            &       76.0                                                             &            92.4                                                        &                     66.7                                               &                                          94.4                         & OR + ALL        
\end{tabular}
\caption{Performance of our gated architecture on the spin classification task. The first line shows the spin task learned on its own, the others show performance when spin classification is paired with other tasks.}
\label{tbl:spin}
\end{table*}

\begin{table*}[]
\centering
\begin{tabular}{l| c c c c c c}
\textbf{Experiment} & \textbf{Spin Acc.} & \textbf{\begin{tabular}[c]{@{}l@{}}Pose \\ AUC@16\end{tabular}} & \textbf{\begin{tabular}[c]{@{}l@{}}Pose \\ AUC@40\end{tabular}} & \textbf{\begin{tabular}[c]{@{}l@{}}Ball \\ AUC@2\end{tabular}} & \textbf{\begin{tabular}[c]{@{}l@{}}Ball \\ AUC@5\end{tabular}} & \textbf{Dataset} \\ \hline
Pose                &    -                &                                             87.4                       &             95.3                                                      & -                                                                  & -                                                                 & ALL               \\
Spin + Pose         &           68.2         &                                             77.2                        &                                                   91.5                 & -                                                                  & -                                                                 & OR               \\
Spin + Pose         &         64.2           &                    83.0                                                 &                  92.3                                                  & -                                                                  & -                                                                 & OR + ALL               \\
Pose + Ball         &  -                  &          83.7                                                          &          92.6                                                         &                   80.9                                                 &                                   95.0                                & ALL         
\end{tabular}
\caption{Performance of our gated architecture on the 2D human pose detection task. The first line shows the pose task learned on its own, the others show performance when pose detection is paired with other tasks.}
\label{tbl:pose}
\end{table*}

We begin with an analysis of the ball detection experiments that we have performed. Table \ref{tbl:architectures} shows our performance on the ball classification task on the three main architectures we tried -- the gated architecture that we have proposed is Section \ref{sec:gated}, LSTMs and a single frame prediction model.

For this experiment, we measure our performance by computing the area under curve (AUC) measurement for precision-recall at 2 and 5 pixels from the ground truth. Each of these curves was carved out by sweeping through the log probability of detections under our model.

We find that, as expected, our simplified gated architecture performs better than alternative architectures.

\subsubsection{LSTM}

For this experiment, we replace the gated architecture used in other experiments with a standard LSTM architecture. This architecture is similar to the LSTM architecture of \cite{lstm}, with the notable exception that all of the fully connected layers have been replaced with convolutional layers. Table \ref{tbl:architectures} shows that the LSTM architecture performs slightly worse for our task than our proposed gated architecture. Part of this may be attributed simply to the difficulty of training the LSTM architecture, since it was much easier to find viable training parameters for our modified gated architecture.

\subsubsection{Single Frame}

This experiment removes the recurrent connections of the previous experiment, replacing the entire recurrent unit with a single convolutional layer. As is easy to imagine, without a persistent state or gating, this system should be unable to learn dependencies or ball dynamics over time. However, we find that it performs remarkably well at 2 pixel distance from ground-truth, outperforming the recurrent model which performs better at 5 pixel threshold. We hypothesize that this is partly attributable to the faster training of the non-recurrent model. It is also possible that the recurrence hurts the precise center computation as the recurrent connections add as a constraint. However, at increased tolerances, the recurrence is able to rule out false positives that the non-recurrent model finds.



\subsection{Spin Classification}

In this section we explore the spin classification task. We perform four main experiments towards this end, shown in Table \ref{tbl:spin}. For all of these experiments, we use the standard gated architecture discussed in Section \ref{sec:gated}. As stated, earlier, all tasks use standard cross-entropy, either over a one-hot vector or a heat map, but performing multi-task learning we multiply the spin classification loss by $3$.

Our first experiment with Spin classification measures results when spin is learned in isolation. The second experiment, wherein pose is learned as an auxiliary task, sees some improvements in spin performance, likely due to the additional understanding of the humans playing.

For the experiments involving both spin and ball tracking, we combine the ``SPIN-OR" and ``SPIN-ALL" dataset splits as described in Section \ref{sec:impl}, while only propagating losses for spin when the data comes from the ``SPIN-OR" dataset. We find here that ball detection does not actually provide any advantage to spin classification. In all likelihood, this is because the signal understanding from the player is far more informative than any signal from the ball.




\subsection{Human Pose}

Although we have introduced the task of human pose primarily with the goal of improving performance in the spin classification task, it nevertheless can be informative to evaluate it as a full-fledged task. We do this in Table \ref{tbl:pose}, using the same top-1 and top-5 precision metrics used previously in Table \ref{tbl:architectures}. Here, we see that pose detection actually performs the best on its own - additional tasks actually cause performance to decline. This isn’t necessarily surprising; our architecture features a very high degree of feature sharing between ball tracking and pose detection, but these tasks are not necessarily related and there is no reason to believe that the two would complement each other. 

In the case of pose and spin learning, while it is natural to expect that pose detection would help with spin classification, it may not be the case that spin classification will help with pose prediction. To simplify the comparison, we perform joint learning of pose and spin using one trial of ``SPIN-OR" and one trial of ``OR + ALL". In addition to finding that spin classification does not particularly help with pose, we see that a lot of our performance can be attributed to the additional data of the ``SPIN-ALL" split, since removing it notably hurts performance.

\section{Conclusion}
We have presented in this paper the new SPIN dataset for the game of ping pong. SPIN is a rich computer vision dataset presenting several problems for computer vision researchers, of which we explore three: pose tracking, spin type prediction, and human pose detection. We present a novel recurrent architecture that provides a strong baseline for these tasks.

{\small
\bibliographystyle{ieee_fullname}
\bibliography{references}
}

\newpage
\section{Supplemental Material}

\noindent\begin{minipage}{\textwidth}
    \centering
    \includegraphics[width=0.9\textwidth]{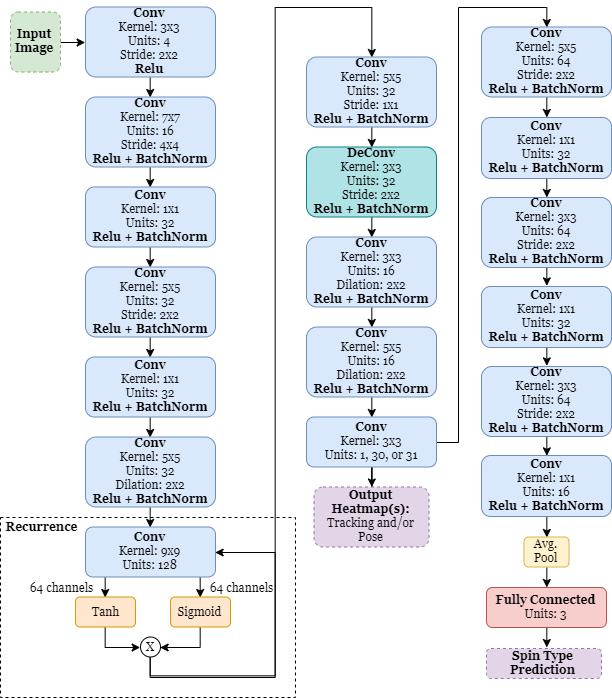}
    \captionof{figure}{The complete architecture of our system, except for the patch prediction model. ``\textbf{Deconv}" refers to the standard transpose convolution operation, ``Dilation" refers to dialated convolutions \cite{Yu2015MultiScaleCA}, and ``\textbf{BatchNorm}" refers to Batch normalization \cite{bnorm}.}
\end{minipage}


\begin{figure*}[!ht]
 \centering
 \includegraphics[width=0.6\textwidth]{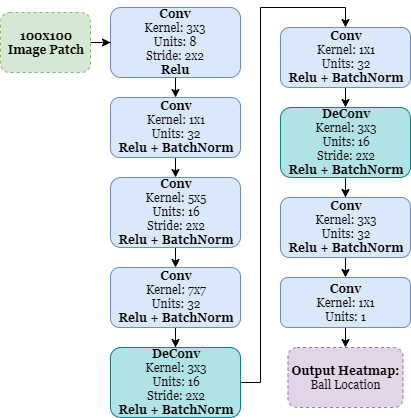}
\caption{The patch prediction model. When performing tracking, after we find the maximum activation in the ball location heatmap, we crop out a $100 \times 100$ patch of the original image centered at that location, and feed it through the above network, creating a new heatmap which we then use for a more precise ball location. ``\textbf{Deconv}" refers to the standard transpose convolution operation, ``Dilation" refers to dialated convolutions \cite{Yu2015MultiScaleCA}, and ``\textbf{BatchNorm}" refers to Batch normalization \cite{bnorm}.}
\label{fig:path_arch}
\end{figure*}

\end{document}